\documentclass[eat,twocolumn]{jmlr}% new name PMLR (Proceedings of Machine Learning Research)

\usepackage{longtable}% for long tables

\usepackage{booktabs}
\usepackage[load-configurations=version-1]{siunitx} % newer version
\theorembodyfont{\upshape}
\theoremheaderfont{\scshape}
\theorempostheader{:}
\theoremsep{\newline}

\jmlrvolume{1}
\firstpageno{1}
\jmlryear{2022}
\jmlrworkshop{Machine Learning for Health (ML4H) 2022}

\usepackage{enumerate}
\usepackage{mathtools}
\usepackage{amsmath,etoolbox}
\usepackage{cases}
\newcounter{tableeqn}[table]
\renewcommand{\thetableeqn}{M.\arabic{tableeqn}}
\newcounter{tablesubeqn}[tableeqn]

\usepackage[font=scriptsize]{caption}
\usepackage{wrapfig}
\newcommand{\indep}{\rotatebox[origin=c]{90}{$\models$}}

\newcommand{\specialcell}[2][c]{%
  \begin{tabular}[#1]{@{}c@{}}#2\end{tabular}}
\title[Clinically Plausible Disentanglement with Structured Variational Priors]{Clinically Plausible Pathology-Anatomy Disentanglement in Patient Brain MRI with Structured Variational Priors}

\author{
  \Name{Anjun Hu}$^{1,2}$, %\Email{anjun@cim.mcgill.ca} \\
  \Name{Jean-Pierre R. Falet}$^{1,2,3}$, % \Email{jpfalet@cim.mcgill.ca}  \\
  \Name{Brennan S. Nichyporuk}$^{1,2}$,  %\Email{nichypob@mila.quebec} \\
  \Name{Changjian Shui}$^{1,2}$,  %\Email{maxshui@cim.mcgill.ca}  \\
  \Name{Douglas L. Arnold}$^{3}$, % \Email{douglas.arnold@mcgill.ca}\\
  \Name{Sotirios A. Tsaftaris}$^{4, 5}$,
  \Name{Tal Arbel}$^{1,2}$ \\ \vspace{.5em}
  \Email{\{anjun, jpfalet, maxshui, arbel\}@cim.mcgill.ca,\\ nichypob@mila.quebec, douglas.arnold@mcgill.ca,  s.tsaftaris@ed.ac.uk}\\ \vspace{.5em}
\addr $^{1}$Centre for Intelligent Machines, McGill University, Montréal, Canada\\
\addr $^{2}$MILA (Québec Artificial Intelligence Institute), Montréal, Canada \\
\addr $^{3}$Brain Imaging Center, Dept. Neurology and Neurosurgery, McGill University, Montréal, Canada \\
\addr $^{4}$School of Engineering, University of Edinburgh, UK \\
\addr $^{5}$The Alan Turing Institute, UK \\ \vspace{-2em}
 }

\begin{document}

\maketitle
\begin{abstract}
  We propose a hierarchically structured variational inference model for accurately disentangling observable evidence of disease (e.g. brain lesions or atrophy) from subject-specific anatomy in brain MRIs.
  With flexible, partially autoregressive priors, our model (1) addresses the subtle and fine-grained dependencies that typically exist between anatomical and pathological generating factors of an MRI to ensure the clinical validity of generated samples; (2) preserves and disentangles finer pathological details pertaining to a patient's disease state.
  Additionally, we experiment with an alternative training configuration where we provide supervision to a subset of latent units. It is shown that (1) a partially supervised latent space achieves a higher degree of disentanglement between evidence of disease and subject-specific anatomy; 
  (2) when the prior is formulated with an autoregressive structure, knowledge from the supervision can propagate to the unsupervised latent units, resulting in more informative latent representations capable of modelling anatomy-pathology interdependencies.

\end{abstract}
\begin{keywords}
Variational Inference, Disentanglement, Medical Image Analysis
\end{keywords}

\section{Introduction}
\label{sec:intro}
In medical image analysis, various representation learning methods have been introduced to disentangle particular anatomical or pathological structures from the the rest of the image under different observations \citep{SotosDisentanglementTutorial, fragemann2022reviewDisentanglement}.
Although disentanglement of observable evidence of disease (e.g. brain lesions or atrophy) from subject-specific anatomy has been shown to be helpful in various downstream tasks \citep{qin2019unsupervisedDisentanglementforRegesteration,DisentangledAAEAgePredictionMissingModality, zhou2021chestCOVIDdisentanglement, zuo2021informationUnsupervisedDomainAdaptationDisentanglement, liu2021semisupervisedDisentanglementdomain-generalised, liu2021unifiedBrainMRTranslationDisentanglement,  D2-NetDisentanglementSegmentationMissingModality}, maintaining the clinical plausibility of the learned representations is particularly challenging in the context of brain MRI analysis, for a number of reasons. 
Firstly, pathological and anatomical features in brain MRIs typically exhibits some degree of dependency, but the exact structure is generally unknown \textit{a priori} or requires extensive domain knowledge to accurately describe. For example, multiple sclerosis (MS) is a chronic neurological disease characterized by T2 hyperintense lesions and gadolinium-enhancing (Gd+) lesions in the brain and spinal cord. MS lesions are typically found in periventricular white matter but cannot occur in certain brain regions (e.g. within the ventricles). 
It is often the case that the observable pathological features in the brain (e.g. hyper-intense MS lesions) cannot be easily disentangled from subject-specific anatomical structures (e.g. sulcal pattern, ventricular shape) in a clinically plausible way due to dependencies between the anatomical and pathological generating factors. 
Secondly, fine-grained pathological features in brain MRIs may have very high clinical significance. 
Although the omission of finer details does not preclude generative models from achieving a good approximation of the overall data distribution (i.e. a good test set log likelihood), such details could be highly meaningful in certain downstream tasks. This is, again, exemplified by MS lesions that can be as small as 3 mm but still represent a significant marker of disease activity \citep{MSLesionSize}. 
A robust and clinically useful representation in the context of brain MRI analysis must therefore: 1) faithfully model the spatial distribution of the lesions and their dependency on patients' brain anatomical structures, and 2) accurately capture and disentangle lesions of all sizes, along with other potential imaging markers characterized by fine-grained details in the images (e.g. white matter texture).

\begin{figure}[h]
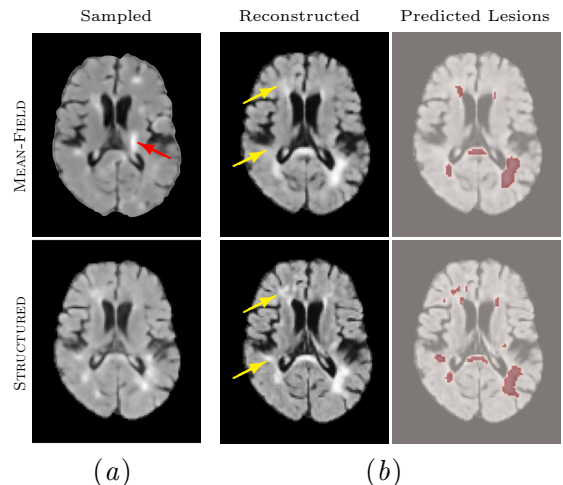

\floatconts
  {fig:IntroFailureModes}
  {\vspace{-2em}\caption{Failure modes of mean-field VAE. (a) A sample drawn from a mean-field model with lesions in clinically invalid locations (red arrow).
  (b) Mean-field model leads to missed small lesions in the reconstructed image (yellow arrows). Our proposed structured model does not suffer from these issues}}
  {
    \subfigure[][b]{\label{fig:IntroInvalidMSLesions}
      {\includeteximage[angle=0]{InvalidLesions}}}
    \subfigure[][b]{\label{fig:IntroBlurredMSLesions}
      {\includeteximage[angle=0]{BlurredLesionsWithArrows}}
    }
  }
\end{figure}

Many existing methods for learning disentangled representations are based on the variational auto-encoder (VAE) \citep{Kingma2014VAE, burgess2018betaVAE, higgins2016betaVAE}. However, a straight adoption of VAE for pathology-anatomy disentanglement in patient brain MRIs is often unsatisfactory due to the aforementioned challenges. There are two common failure modes. Firstly, mean-field variational inference poses the unlikely assumption that all generating factors are independent \citep{VariationalReview}, thereby failing to capture the inherent dependencies that exist between the anatomical and pathological generative factors in patient brain images. This may result in the synthesis of clinically implausible samples, as depicted in \figureref{fig:IntroInvalidMSLesions}, where lesions appear in clinically impossible regions (red arrow). 
Moreover, the independence assumption may exacerbates another pitfall of VAEs \citep{VariationalReview}, namely, the tendency to suffer from latent mode-covering or over-generalisation \citep{hu2018onUnifying}. 
This may lead to lower synthesis quality or an inability to preserve crucial finer details in medical images. \figureref{fig:IntroBlurredMSLesions} depicts such behaviour in a mean-field model where small lesions are obscured in the reconstructed images. Failure to capture fine details in the learned representation could lead to significantly poorer performance in downstream tasks.

In this work, we propose to address these issues by using structured variational inference \citep{StructuredVI} for fine-grained pathology-anatomy disentanglement in brain MRI. Specifically we model the dependencies that typically exist between pathological and anatomical features via multi-scale VAEs with a hierarchical latent structure (\figureref{fig:GraphicalModel}). We evaluate the effect of different priors in the reconstruction quality and the degree of disentanglement both quantitatively and qualitatively. We find that more expressive structured priors indeed lead to higher reconstruction quality and the preservation of important small pathological details. We verify that the model is capable of pathology disentanglement in an unsupervised setting. With an optional supervision objective, the model is shown to achieve a higher degree of disentanglement and to  be capable of capturing latent dependency.

\section{Model}
\label{sec:Model}
Our model accepts image space observations sampled from a dataset $\mathbf{x} \in \mathbb{R}^{w_\mathbf{x} \times h_\mathbf{x}} \sim \mathcal{D}$ as inputs. 
The model adheres to the customary setup of VAEs with a structured latent space consisting of $(L+1)$ disjoint variable groups (layers) that follow a hierarchical structure, as portrayed in \figureref{fig:GraphicalModel}. Each group or layer consists of spatial latent variables \citep{SpatialGMVAE} at various resolutions scales, denoted as $\mathbf{z} = \{ \mathbf{z}_l \in \mathbb{R}^{w_l \times h_l \times c_l }; l \in \{0, 1, ..., L\} \}$. The inference and generative models can therefore be expressed as follows: 
\begin{equation}
\begin{gathered}
    \label{eq:jointdistributions}
     q_{\phi, \theta}(\mathbf{z}|\mathbf{x}) := q_{\phi_L, \theta_L}(\mathbf{z}_L|\mathbf{x}) 
     \prod_{l=0}^{L-1} q_{\phi_l, \theta_l}(\mathbf{z}_{l}|\mathbf{z}_{l+1})\\ 
     p_\theta(\mathbf{x}, \mathbf{z}) := 
    p_{\theta_x}(\mathbf{x}|\mathbf{z}_L)  p(\mathbf{z}_0) \prod_{l=1}^L p_{\theta_l}(\mathbf{z}_{l}|\mathbf{z}_{l-1})
\end{gathered}
\end{equation}

\begin{figure}[h]
\floatconts
{fig:GraphicalModel}
{\vspace{-2em}\caption{The hierarchically structured latent space with $(L+1)$ disjoint variable groups (layers), encoder distribution parameters ($\Delta \mu_l, \Delta \sigma_l$) and decoder distribution parameters ($\mu_l, \sigma_l$). Dashed lines are active during inference. Solid lines are active during generation and, if residual parameterisation is used, also during bidirectional inference. 
}}
{\includegraphics[width=\linewidth]{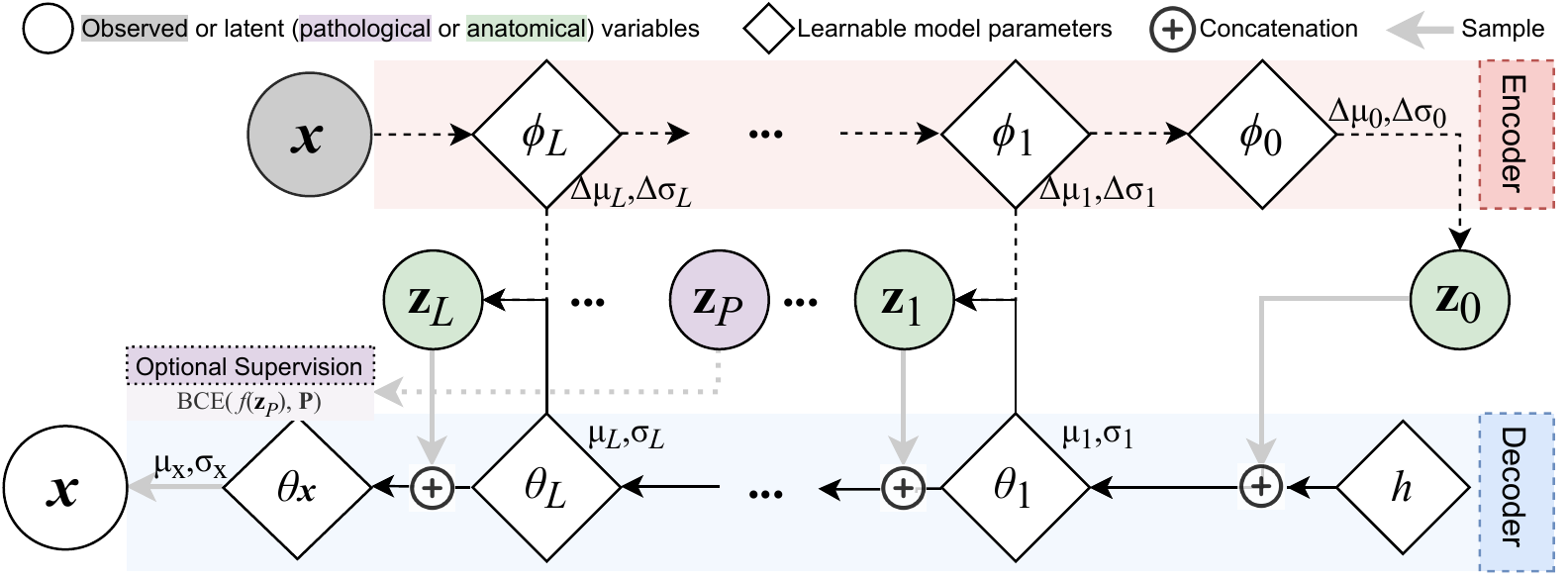}}
\end{figure}

\begin{figure*}[t]
\floatconts
{fig:ArchDiagram}
{\vspace{-2em}\caption{Network Architecture}}
{\includegraphics[width=\linewidth]{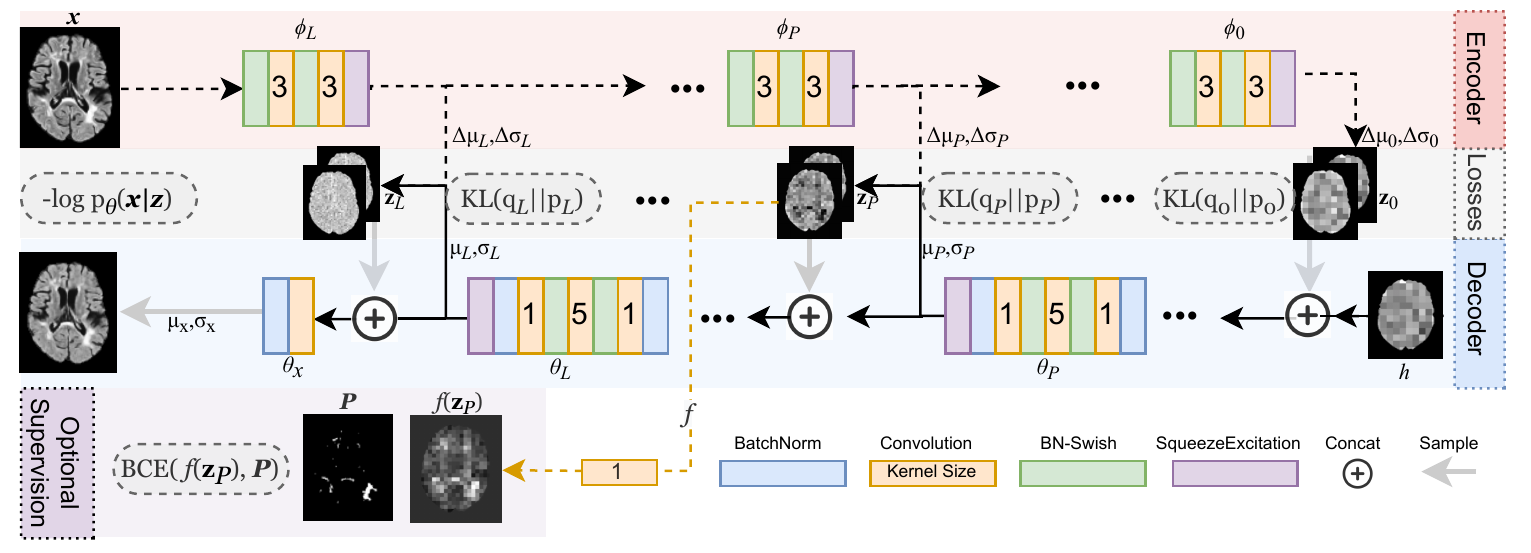}}
\end{figure*}

\begin{table*}[htbp]
 % The first argument is the label.
 % The caption goes in the second argument, and the table contents
 % go in the third argument.
\floatconts
  {tab:ModelParametrizations}%
  {\caption{ELBOs for all four model parameterisation options examined in this work. 
}}%
  { \fontsize{9.2pt}{5pt} \vspace{-2em}
     \centering
    \begin{tabular}{c|c}
    Model & ELBO \\
    \hline \hline
    %  \eqref{eq:BaselineELBO} 
    \specialcell{\textsc{VAE} \\
    \refstepcounter{tableeqn} (\thetableeqn)
    \label{eq:BaselineELBO}
    }
    &
    $\begin{array} {r@{}l@{}}
    \mathbb{E}_{\mathbf{x} \sim \mathcal{D}} 
    \Bigl[
    \mathbb{E}_{\mathbf{z} \sim q_{\phi}(\mathbf{z}|\mathbf{x})}  
    [\log p_\theta( \mathbf{x}|\mathbf{z})] - 
    \sum_{l=0}^{L} \mathrm{KLD}[q_{\phi}
    (\mathbf{z}_l|\mathbf{x}) || \mathcal{N} (\mathbf{0}, \mathbf{I})  ]
    \Bigr] 
    \end{array}$
    
    \\
    
    %  \eqref{eq:NVAE-ELBO} 
    \specialcell{\textsc{nVAE}\\
    \refstepcounter{tableeqn} (\thetableeqn)
    \label{eq:NVAE-ELBO}
    }
    &
    $\begin{array} {r@{}l@{}}
    \mathbb{E}_{\mathbf{x} \sim \mathcal{D}} 
    \Bigl[
    \mathbb{E}_{\mathbf{z} \sim q_{\phi, \theta}(\mathbf{z}|\mathbf{x})}  
     [
    \log p_\theta( \mathbf{x}|\mathbf{z})
    ] 
     - \mathrm{KLD}[q_{\phi}(\mathbf{z}_0| \mathbf{x})||\mathcal{N} (\mathbf{0}, \mathbf{I})]
    \\ -   
    \sum_{l=1}^{L} 
    \mathbb{E}_{q_{\phi, \theta} (\mathbf{z}_{<l} | \mathbf{x})}
    \mathrm{KLD}[q_{\phi, \theta}(\mathbf{z}_l|\mathbf{z}_{<l}, \mathbf{x})||p_\theta(\mathbf{z}_l | \mathbf{z}_{<l})] 
    \Bigr] 
    \end{array}$
    \\
    
    \specialcell{\textsc{nVMP} \\
    % \begin{tabular}{C{1.2cm}|C{1.2cm}}  \checkmark & \checkmark \end{tabular}\\
    \refstepcounter{tableeqn} (\thetableeqn)
    \label{eq:VampNVAE-Z0-ELBO}
    }

    &
    $\begin{array} {r@{}l@{}}
    \mathbb{E}_{\mathbf{x} \sim \mathcal{D}} \Biggl[  
    \mathbb{E}_{\mathbf{z} \sim q_{\phi, \theta}(\mathbf{z}|\mathbf{x})}  
     [\log p_\theta( \mathbf{x}|\mathbf{z})]
     - \mathrm{KLD} \bigl[ q_{\phi}(\mathbf{z}_0| \mathbf{x})||
    \frac{1}{K}\sum_{k=1}^K q_{\phi}(\mathbf{z}_0| \mathbf{u}_k) \bigr]
    \\ -   
    \sum_{l=1}^{L} 
    \mathbb{E}_{q_{\phi, \theta} (\mathbf{z}_{<l} | \mathbf{x})}
    \mathrm{KLD}[q_{\phi, \theta}(\mathbf{z}_l|\mathbf{z}_{<l}, \mathbf{x})||p_\theta(\mathbf{z}_l | \mathbf{z}_{<l})] 
    \Biggr] 
    \end{array}$
    \\
    
    \specialcell{\textsc{nVMP+} \\
    \refstepcounter{tableeqn} (\thetableeqn)
    \label{eq:VampNVAE-All-ELBO}
    }
    &
    $\begin{array} {r@{}l@{}}
    \mathbb{E}_{\mathbf{x} \sim \mathcal{D}} \Biggl[  
    \mathbb{E}_{\mathbf{z} \sim q_{\phi, \theta}(\mathbf{z}|\mathbf{x})}  
     [\log p_\theta( \mathbf{x}|\mathbf{z})]
     - \mathrm{KLD} \bigl[ q_{\phi}(\mathbf{z}_0| \mathbf{x})||
    \frac{1}{K}\sum_{k=1}^K q_{\phi}(\mathbf{z}_0| \mathbf{u}_k) \bigr]
    \\ -   
    \sum_{l=1}^{L} 
    \mathbb{E}_{q_{\phi, \theta} (\mathbf{z}_{<l} | \mathbf{x})}
    \biggl[ 
    \mathrm{KLD}[q_{\phi, \theta}(\mathbf{z}_l|\mathbf{z}_{<l}, \mathbf{x})||p_\theta(\mathbf{z}_l | \mathbf{z}_{<l})] 
    + \mathrm{KLD}[p_\theta(\mathbf{z}_l | \mathbf{z}_{<l}) || 
    \frac{1}{K}\sum_{k=1}^K q_{\phi}(\mathbf{z}_l| \mathbf{u}_k)] 
    \biggr]
    \Biggr] 
    \end{array}$ 
  \end{tabular}
  }
\end{table*}

In this paper, we examine four ways to construct the ELBO objective, summerized in \tableref{tab:ModelParametrizations}.

\vspace{0.5em}
\noindent
\textbf{(1)} VAE \eqref{eq:BaselineELBO}: a vanilla multi-scale \textsc{VAE} with mean-field, parameter-free standard Gaussian priors at each layer. This model has a ``hierarchy'' in the sense of having latent representations at various resolution scales, but is not ``hierarchical'' in its distributional parametrization as there is no explicit inter-group dependency nor explicit information sharing between the encoder and the decoder. In this parametrization:
\begin{subequations}
\setlength{\jot}{-1pt}
    \begin{align}
    \begin{split}
    \label{eq:vanillaposteriors}
    & q_{\phi_l}(\mathbf{z}_l) := \mathcal{N} (\Delta \mu_{\phi_{>l}} (\mathbf{x}), \Delta \sigma_{\phi_{>l}}( \mathbf{x}) )
    \end{split} 
    \end{align} \\[-40pt]
    \begin{align}
    \label{eq:vanillapriors} 
    & p_{\theta_l}(\mathbf{z}_l) := \mathcal{N} (\mathbf{0}, \mathbf{I}) \;\;
    \forall l \in \{0, 1, ..., L\}
    \end{align} 
\end{subequations}

\vspace{0.5em}
\noindent
\textbf{(2)} \textsc{nVAE} \eqref{eq:NVAE-ELBO}: a hierarchical model with residual normal parameterisation proposed by \cite{NVAE} and \cite{LadderRasmusVHBR15}. The features that set this model apart from its vanilla counterpart are the explicit information sharing between the encoder and the decoder networks, as well as its partially auto-regressive nature.
Firstly, unlike conventional VAEs, decoder parameters $\theta$ in \textsc{nVAE} not only characterize the generative distribution $p_\theta$ \eqref{eq:relativevariationalpriors} but are also a part of the inference model and hence play an important role in characterizing the posterior distribution $q_{\phi, \theta}$ \eqref{eq:relativevariationalposteriors}. For latent groups other than the topmost one $l>0$, the inference model is \textit{bidirectional}. It estimates the \textit{relative variational} posteriors \eqref{eq:relativevariationalposteriors} that characterize the deviation from priors obtained from preceeding layers of the decoder. With this design, KL optimization is expected to be simpler than when posteriors predict the absolute mean and variances at each layer.
\begin{subequations}
\begin{multline}
    \label{eq:relativevariationalposteriors}
    q_{\phi, \theta}(\mathbf{z}_l|\mathbf{z}_{<l}, \mathbf{x}) :=
    \mathcal{N} 
    \Bigl(
    \mu_{\theta_{<l}}(\mathbf{z}_{<l}) + \Delta \mu_{\phi_{>l}}(\mathbf{x}), \\[-5pt] \sigma_{\theta_{<l}}(\mathbf{z}_{<l}) \cdot \Delta \sigma_{\phi_{>l}}(\mathbf{x}) 
    \Bigr)
\end{multline}\\[-40pt]
\begin{align}
    \label{eq:relativevariationalpriors}
    p_\theta(\mathbf{z}_l|\mathbf{z}_{<l}) := \mathcal{N} (\mu_{\theta_{<l}}(\mathbf{z}_{<l}), \sigma_{\theta_{<l}}(\mathbf{z}_{<l}))
\end{align}
\end{subequations}

Furthermore, \textsc{nVAE} is considered to be partially auto-regressive and hence a more expressive prior than the standard mean-field parametrization. While the prior for each group $p(\mathbf{z}_l)$ is dependent on those of the preceding layers $p(\mathbf{z}_{<l})$, each element within the same latent group $\mathbf{z}_l := \{z^i_l, ...\}$ still adhere to the independence assumption as $ \forall l \in \{0, 1, ..., L\}, z^i_l \indep z^j_l;  i \neq j$. 
We can hence calculate the relative KLD loss for each element with a simple analytic expression:
\begin{multline}
\label{eqn:RKLDdef}
    \mathrm{(Relative)\;KLD}\;[q_{\phi, \theta}(z^i_l|\mathbf{x})||p_{\theta}(z^i_l)] = \\
    \frac{1}{2} \Biggl( 
    \frac{(\Delta \mu_l^i)^2}{(\sigma_l^i)^2} + (\Delta \sigma_l^i)^2 - \log (\sigma_l^i)^2 -1 
    \Biggr) 
\end{multline}

We propose two extensions to \textsc{nVAE} by integrating a VamPrior \citep{VamPriorTomczakW17} into the hierarchical VAE setup for extra flexibility and expressiveness in the hierarchical latent structure. 
By incorporating encoder parameters in the trainable prior, we expect the following two models to achieve a greater extent of  ``coupling'' or ``collaboration'' between the priors and the posteriors.

\vspace{0.5em}
\noindent
\textbf{(3)} \textsc{nVMP} \eqref{eq:VampNVAE-Z0-ELBO} replaces the standard-Gaussian prior of the topmost layer of \textsc{nVAE} with a $K$-component multimodal VamPrior $\frac{1}{K}\sum_{k=1}^K q_{\phi}(\mathbf{z}_l| \mathbf{u}_k)$  characterized by trainable pseudo-inputs $\mathbf{u}$ and encoder parameters $\phi$. 
The subsequent layers still adhere to the hierarchical residual parameterisation of \textsc{nVAE} (we retain the residual Gaussian parameterisation for subsequent layers). The implication is that only $\mathbf{z}_0$ is multi-modal whereas the distributions of all lower level ``residual deviations'' are assumed to be Gaussian. 

\vspace{0.5em}
\noindent
\textbf{(4)} \textsc{nVMP+} \eqref{eq:VampNVAE-All-ELBO} extends \textsc{nVMP} with one more KL term between the encoder-driven VamPriors and the decoder priors is imposed for the entire hierarchy. In this case, the decoder ``priors'' are regarded as ``intermediate posteriors'' and encouraged to imitate the encoder-driven multi-modal distribution throughout the hierarchy. We postulate that this configuration adds an extra layer of information sharing between the encoder and the decoder networks which can potentially lead to further improvement in representation quality. 

\section{Experiments and Results}
\label{sec:ExperimentsResults}
% Dataset
We validate our approach on two brain MRI datasets: the publically available Alzheimer's Disease Neuroimaging Initiative (ADNI) dataset \citep{mueller2005adni} ($N=864$), and a proprietary MS dataset from a MS clinical trial ($N=815$). The central 16 2-D slices of T1-weighted sequences were used for the AD experiments, while the central 24 2-D slices of Fluid Attenuated Inverse Recovery (FLAIR) sequence were used for the MS experiments. Expert T2 lesion segmentation labels for the MS experiments were provided. Both datasets were divided into non-overlapping training (60 \%), validation (20 \%) and testing (20 \%) sets. Additional acquisition and pre-processing details are described in Appendix~\ref{apd:Data}.

We first train the model under an unsurpervised setting and evaluate the effect of incorporating additional prior structures on synthesis quality. 
As shown in Table \ref{tab:MainResults}, VAEs with more expressive structured priors indeed outperform their mean-field counterpart at the same model capacity in terms of image reconstruction fidelity.

\begin{table}[h]
\floatconts
 {tab:MainResults}
 {\caption{\centering Reconstruction quality metrics. \protect\linebreak \citep{wang2004SSIM, FID}}}
 { \centering \small \vspace{-2em}
  \subtable[MS]{%
    \begin{tabular}{c|cccc}
        \hline
Model  &  LL$\uparrow$  & PSNR$\uparrow$   & SSIM$\uparrow$  &   FID$\downarrow$  \\
\hline \hline
\textsc{VAE}  & 2758 & 25.1 & 0.72 & 0.058  \\
\textsc{nVAE}  & 2458 &  25.7 & 0.75 & \textbf{0.023}  \\
\textsc{nVMP}  & 2374 & 25.9 & 0.75 & 0.031 \\
\textsc{nVMP+}  & \textbf{1953} &  \textbf{26.6} & \textbf{0.79} & 0.035 
     \end{tabular}
  }
  \subtable[AD]{%
    \begin{tabular}{c|cccc}
    \hline
    Model            &  LL$\uparrow$  & PSNR$\uparrow$   & SSIM$\uparrow$  &   FID$\downarrow$  \\
  \hline \hline
  \textsc{VAE} & 2386 & 24.6  & 0.70 & 0.030 \\
  \textsc{nVAE} & 2105 & 25.2  & 0.73 & 0.013  \\
  \textsc{nVMP} & 1863 & 25.5  & 0.75 & 0.011 \\
  \textsc{nVMP+}& \textbf{842} & \textbf{26.8}  & \textbf{0.80} & \textbf{0.007} 
    \end{tabular}
  }
 }
\end{table}

We additionally examine model behaviours in a supervised learning setting depicted in the bottom-left (purple) block in \figureref{fig:GraphicalModel} and \figureref{fig:ArchDiagram}. In this setting, we supplement the MS model with a lesion segmentation objective between a chosen ``pathological'' latent subset $\mathbf{z}_P$ and expert pathology (lesion) segmentation labels $\mathbf{P}$. The rest of the latent space (that remains unsupervised) are regarded as the anatomical latent subsets, denoted as $\mathbf{z}_A$.

Firstly, as one might expect, supervision is shown to enhance latent disentanglement as one may anticipate.
Disease-related features in the synthesized images are noticeably more \textit{sensitive} \citep{VAEPCA} to perturbations in $\mathbf{z}_P$ comapred to $\mathbf{z}_A$, as shown in \figureref{fig:AttributeSensitivity}, Appendix~\ref{apd:AdditionalVisuals}. Such a disparity in attribute sensitivity is appreciable in unsupervised models, but is made much more pronounced by the selective latent supervision.

Secondly and more importantly, \emph{supervision helps to verify that the model is indeed actively using the latent structures.} In models with autoregressive structures (\textsc{nVAE}, \textsc{nVMP}, \textsc{nVMP+}), knowledge from the supervision is propagated to the unsupervised ``anatomical'' latent units $\mathbf{z}_A$, as in, those unsupervised latent units attain a higher linear predictability (Lasso regression $R^2$ scores 
\citep{eastwood2018frameworkDisentanglement}, Table \ref{tab:Informativeness}) with respect to lesion volume. This is in contrast to the behaviour of the baseline mean-field VAE, where information from the supervision task is constrained within the supervised group $\mathbf{z}_P$. This observation shows that the model is indeed taking advantage of the extra structures brought by the autoregressive priors and the residual parameterisation and hence, indeed capable of modelling the dependencies between anatomical and pathological generating factors.

\begin{figure}[h]
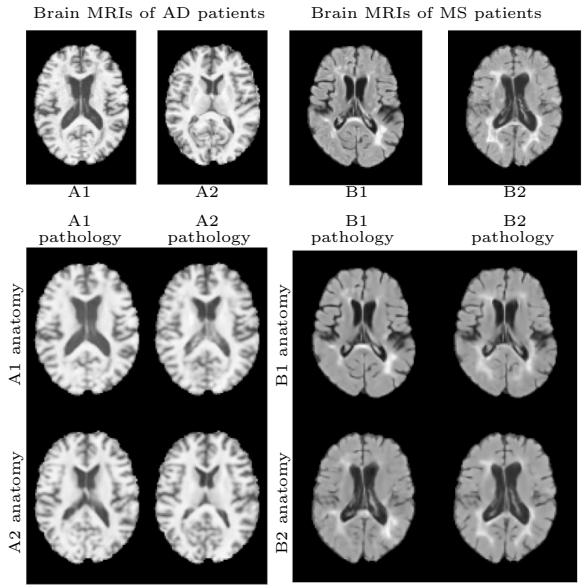

\floatconts
  {fig:StyleMixing}
  {\vspace{-1.7em}\caption{Pathology-anatomy disentanglement visualised by ``style-mixing'' between pairs of Alzheimer's Disease (AD) and Multiple Sclerosis (MS) patient brain MRIs.}}
  {\includeteximage[angle=0]{StyleMixing} }
\end{figure}
We can qualitatively evaluate pathology-anatomy disentanglement by swapping anatomical and pathological latent features between a pair of subjects in a manner similar to ``style-mixing'' \citep{StyleGAN}. As shown in \figureref{fig:StyleMixing} for representative examples, brain atrophy in AD patients (left), and T2 lesions in MS patients (right), are disentangled from the subject's anatomical particularities (such as sulcal pattern), thus enabling the mixing the pathology of one patient with the anatomy of the other. 

We may also leverage conditional distributions learned by the model to examine subject-specific pathology distributions. For example, based on learned representations of Subject B1 in \figureref{fig:StyleMixing}, we may visualise many possible disease states given this subject's anatomy (\figureref{fig:ConditionalResampling}, top row) or explore how this subject's lesions would manifest on other subjects’ brain anatomies (\figureref{fig:ConditionalResampling}, bottom row).

\begin{figure}[t]
\floatconts
  {fig:ConditionalResampling}
  {\vspace{-1.7em}\caption{Visualising conditional distributions. Images in the top row are generated by fixing $\mathbf{z}_A$ to that of Subject B1 and resampling the layer corresponding to $\mathbf{z}_P$. Images in the bottom row are generated by obtaining $\mathbf{z}_A$ from other real samples and fixing $\mathbf{z}_P$ to that of Subject B1.}
  }
  {
     \hspace{-0.8pt}\includegraphics[clip, trim={0.cm 0.3cm 6.5cm 0.3cm}, width=\linewidth]{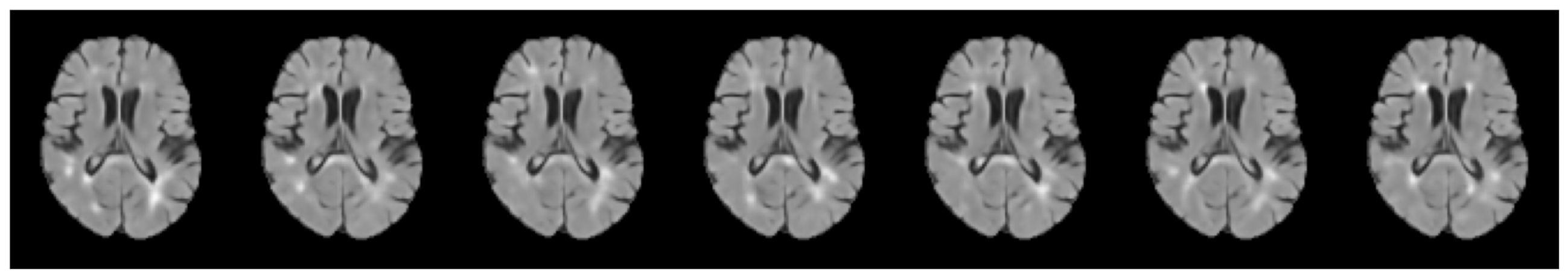}
     \includegraphics[clip, trim={0.cm 0.3cm 6.5cm 0.3cm}, width=.992\linewidth]{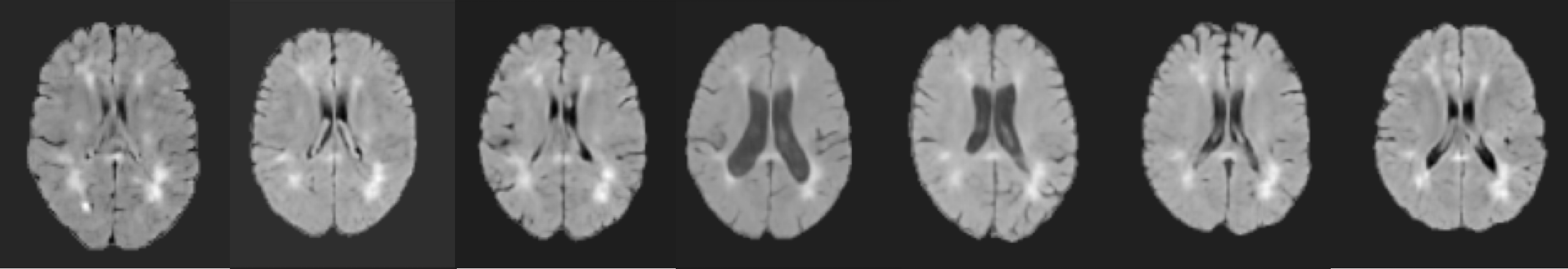}
  }
\end{figure}

\section{Conclusions}
\label{sec:Conclusions}
We propose hierarchical VAEs with structured priors for learning pathology-anatomy disentangled representations of brain MRIs. Our model can faithfully capture imaging features, including fine-grained details, while accounting for pathology-anatomy dependencies to ensure sample validity.
We additionally examine model bevaviours in a supervised learning setting. Supervision is shown to (1) further enhance latent disentanglement; and (2) enable the inspection of information propagation between latent groups for modelling pathology-anatomy interdependencies.
Our model allows for robust and controllable brain MRI synthesis rich in high-frequency and pathologically-sound details, which could be meaningful for various downstream tasks. 

\newpage
\acks{The authors are grateful to the International Progressive MS Alliance 
for supporting this work (grant number: PA-1412-02420), and to the 
companies who generously provided the clinical trial data that made it 
possible: Biogen, BioMS, MedDay, Novartis, Roche / Genentech, and Teva.
Funding was also provided by the Natural Sciences and Engineering Research Council of Canada, the Canadian Institute for Advanced Research (CIFAR) Artificial Intelligence Chairs program, and a technology transfer grant from Mila - Quebec AI Institute. S.A.\ Tsaftaris acknowledges the support of Canon Medical and the Royal Academy of Engineering and the Research Chairs and Senior Research Fellowships scheme (grant RCSRF1819 /\ 8 /\ 25). Supplementary computational resources and technical support were provided by Calcul Québec and the Digital Research Alliance of Canada. Falet, J.-P., was supported by an end MS Personnel Award from the Multiple Sclerosis Society of Canada, by a Canada Graduate Scholarship-Masters Award from the Canadian Institutes of Health Research, and by the Fonds de recherche du Québec - Santé / Ministère de la Santé et des Services sociaux training program for specialty medicine residents with an interest in pursuing a research career, Phase 1. This work was made possible by the end-to-end deep learning experimental pipeline developed in collaboration with our colleagues Justin Szeto, Eric Zimmerman, and Kirill Vasilevski. Additionally, the authors would like to thank Louis Collins and Mahsa Dadar for preprocessing the MRI data.}

\bibliography{Paper106}

\newpage

\appendix

\section{Data Acquisition, Implementation and Training Details}\label{apd:Implementation}\label{apd:Data}
All MRI sequences were acquired at a resolution of 1 mm $\times$ 1 mm $\times$ 1 mm, Each 2-D slice was downsampled to a resolution of 2 mm $\times$ 2 mm. These were standardized to have zero-mean and unit variance. 

We compare the four parameterisations in Table \tableref{tab:ModelParametrizations} with a 5-layer model ($L=4$) the exact same capacity. For each dataset, the latent space capacity is set to 
$ \{
\mathbf{z}_L \in \mathbb{R}^{w_\mathbf{x} \times h_\mathbf{x} \times 2},
\mathbf{z}_{L-1} \in \mathbb{R}^{(w_\mathbf{x}/2) \times (h_\mathbf{x}/2) \times 2}, ... ,
\mathbf{z}_{0} \in \mathbb{R}^{(w_\mathbf{x}/2^L) \times (h_\mathbf{x}/2^L) \times 2} 
\}$.
We use the Adam optimizer \citep{kingma2014adam} with a learning rate of 5e-5 and s weight decay of 1e-8. 

Two loss re-weighting mechanisms are used in our training procedure:
(1) We use a linear annealing schedule \citep{fu_cyclical_2019_naacl} for KLD losses with a cycle length of 10000 iterations. The initial KLD learning rate is set to 2e-7.
(2) To avoid posterior collapse, we use a KL Balancing trick suggested by \citep{NVAE}. 
We re-scale each KL term of the hierarchy with a coefficient proportional to the size of each latent layer as well as the KLD value of that layer. This mechanism encourages more balanced information attribution to each latent layer \citep{vahdat2018dvae++LayerScalingKL, chen2016variationalLayerScalingKL}.

\section{Additional Results}
\label{apd:AdditionalVisuals}

As discussed in Section \ref{sec:ExperimentsResults}, we evaluate layer-wise latent pathology informativeness of MS models by examining each layer's linear predictability of a salient pathological attribute, T2 lesion volume. To quantify linear predictability, we train Lasso regressors ($\alpha=10$) with latent representations obtained from each individual latent layer of each model and compute each Lasso regressors's $R^2$ scores with respect to T2 lesion volume based on expert segmentation labels. 

\begin{table}[h]
\floatconts
 {tab:Informativeness}
 {\caption{(MS) Layer-wise latent informativeness with respect to T2 lesion volume. Models with prefix ``ps-'' have partially supervised latent spaces (i.e. are $\mathbf{z}_P$-supervised). }}
 { \centering \small \vspace{-2em}
\begin{tabular}{c|ccccc}
\hline
Model & $\mathbf{z}_0$ & $\mathbf{z}_1$ & $\mathbf{z}_P$ & $\mathbf{z}_3$ & $\mathbf{z}_4$ \\
\hline \hline
\textsc{VAE} & 0.20 & 0.08 & 0.28 & 0.01 & 0.00 \\
\textsc{nVAE} & 0.11 & 0.12 & 0.00 & 0.00 & 0.00 \\
 \textsc{nVMP} & 0.06 & 0.23 & 0.01 & 0.01 & 0.00 \\
\textsc{nVMP+} & 0.03 & 0.00 & 0.20 & 0.00  & 0.00 \\
\hline
ps\textsc{VAE}  & 0.00& 0.00 &  0.62 & 0.08 & 0.01 \\
ps\textsc{NVAE} & 0.31 & 0.20 & 0.65 & 0.02 & 0.00 \\
ps\textsc{NVMP} & 0.54& 0.23 & 0.63 & 0.02  & 0.01 \\
ps\textsc{NVMP}+ & 0.21 & 0.37 & 0.56 & 0.09 & 0.00 \\
\end{tabular} }
\end{table}

In \tableref{tab:Informativeness}, rows 1-4 show the $R^2$ scores of the unsupervised models, which are generally poor; rows 5-8 show the same metrics for supervised models where supervision is provided to $\mathbf{z}_2$ ($\mathbf{z}_P$) as an additional lesion segmentation objective. Models with autoregressive structures (\textsc{nVAE}, \textsc{nVMP}, \textsc{nVMP+}) benefit more from the supervision - knowledge from the supervision is propagated to the unsupervised ``anatomical'' latent units, resulting in higher $R^2$ scores even in the unsupervised latent subsets. This shows that the model is indeed actively using the latent structures.

Furthermore, by separately scaling each latent subgroup and seeing the changes in the generated images, we can examine the features captured by each individual latent group. Latent disentanglement, as indicated by the remarkable disparity in layer-wise pathological attribute sensitivity to scaling, is made evident with such visualisation.
\begin{figure}[h]
\floatconts
  {fig:AttributeSensitivity}
  {\vspace{-2em}\caption{(MS) Layer-wise pathological attribute sensitivity visualised by individually scaling each layer in the latent hierarchy, from $\mathbf{z}_0$ (top row) to $\mathbf{z}_4$ (bottom row).} 
  \vspace{-1em}
  }
  {
        \includegraphics[clip, trim={4.5cm 0.2cm 4.5cm 0.2cm}, width=\linewidth]{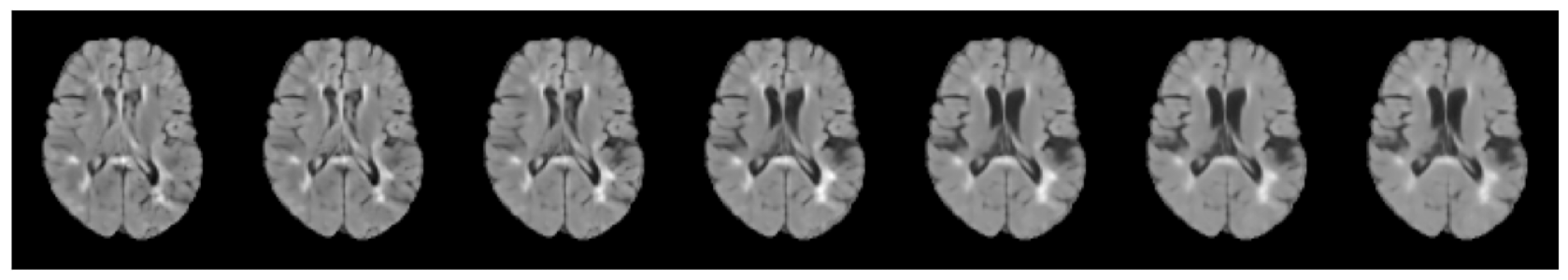}
         \includegraphics[clip, trim={4.5cm 0.2cm 4.5cm 0.2cm}, width=\linewidth]{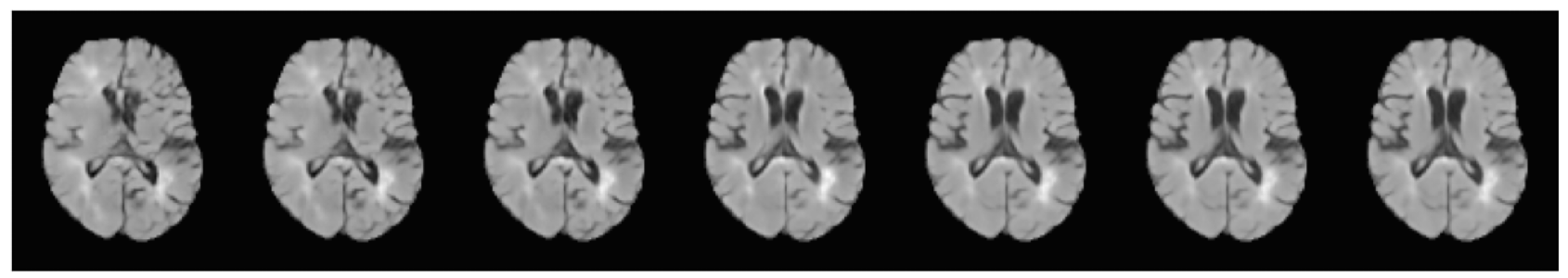}
         \includegraphics[clip, trim={5cm 0.2cm 5cm 0.2cm}, width=\linewidth]{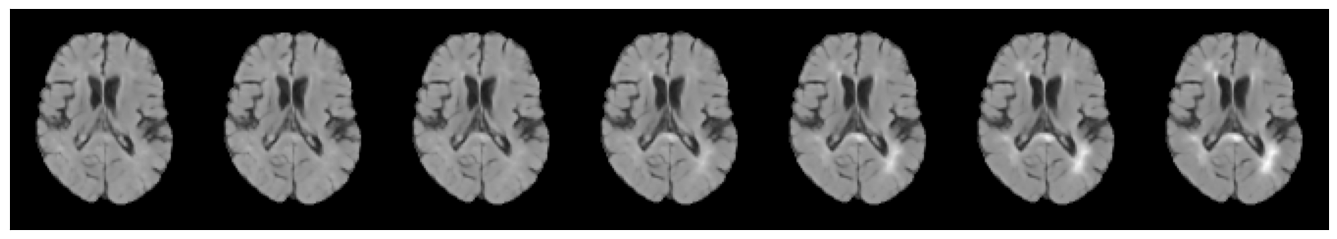}
         \includegraphics[clip, trim={5cm 0.2cm 5cm 0.2cm}, width=\linewidth]{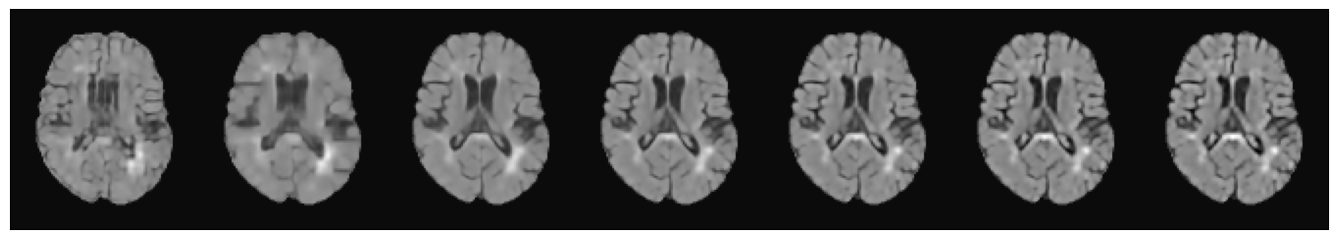}
         \includegraphics[clip, trim={5cm 0.2cm 5cm 0.2cm}, width=\linewidth]{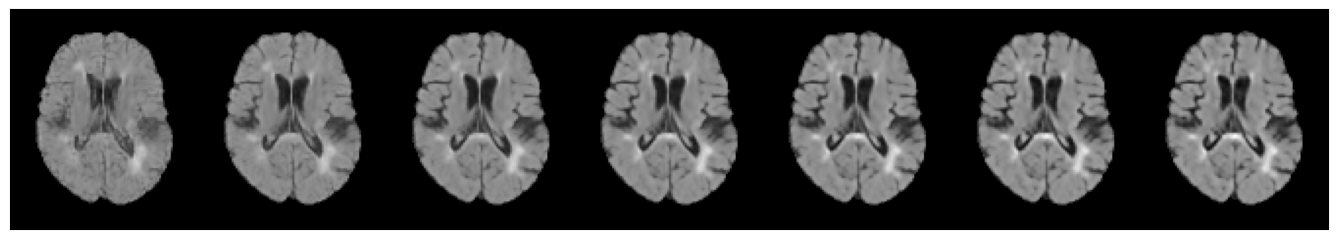}
         \vspace{-1em}
         \begin{picture}(1,1)
            \put(-115,242){\textcolor[HTML]{67AB9F}{\tiny $\mathbf{z}_0$}}
            \put(-115,190){\textcolor[HTML]{67AB9F}{\tiny $\mathbf{z}_1$}}
            \put(-117,142){\textcolor[HTML]{A680B8}{\tiny $\mathbf{z}_2$}}
            \put(-120,137){\textcolor[HTML]{A680B8}{\tiny $\left(\mathbf{z}_P\right)$}}
            \put(-115,87){\textcolor[HTML]{67AB9F}{\tiny $\mathbf{z}_3$}}
            \put(-115,40){\textcolor[HTML]{67AB9F}{\tiny $\mathbf{z}_4$}}
          \end{picture}
  }
\end{figure}
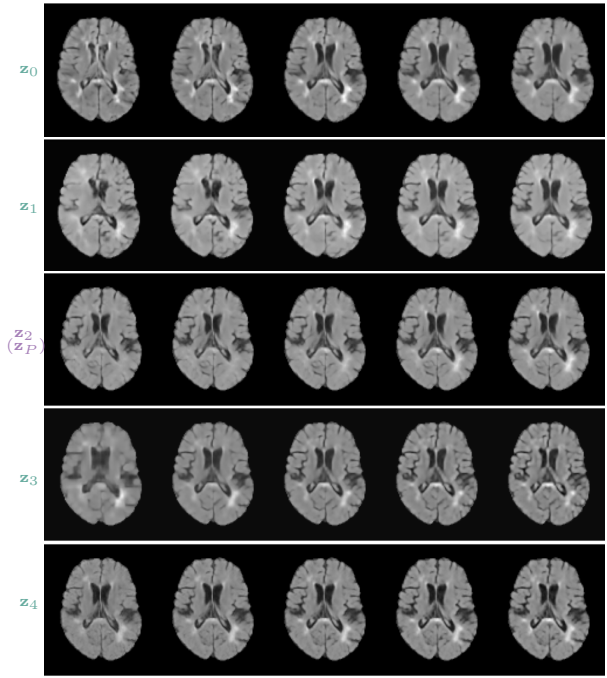

In this particular example (\figureref{fig:AttributeSensitivity}), the appearance of the hyper-intense MS lesions in the synthesised images is relatively insensitive to multiplicative perturbation in all but one latent layer, $\mathbf{z}_2$. The layer with the highest pathological attribute sensitivity, $\mathbf{z}_2$, is hence considered to be a disentangled ``pathological'' latent subset $\mathbf{z}_P$.\\
We note that even in the unsupervised setting, disease-related features in the synthesized images are noticeably more  sensitive to changes in a small subset of latent variables than the rest, which allows us to identify such a subset as $\mathbf{z}_P$ and the rest as $\mathbf{z}_A$ (anatomical latent subsets) in a post-hoc manner. Such disparity in pathological attribute sensitivity is much more pronounced in the ``selective supervision'' setting (bottom-left purple block in \figureref{fig:GraphicalModel} and  \figureref{fig:ArchDiagram}), where the additional supervision is given to a chosen layer $\mathbf{z}_P$.

\begin{figure}[p]
\floatconts
  {fig:VariationPerLayerADNI}
  {\vspace{-2em}\caption{(AD) Variations captured by each layer of the model. Images at the top row are fully resampled at each level of the hierarchy. On each subsequent row $n$, we show the residual variation of layer $n$ by fixing latent codes at the top $(n-1)$ layers. } }
  {
    \includegraphics[width=\linewidth]{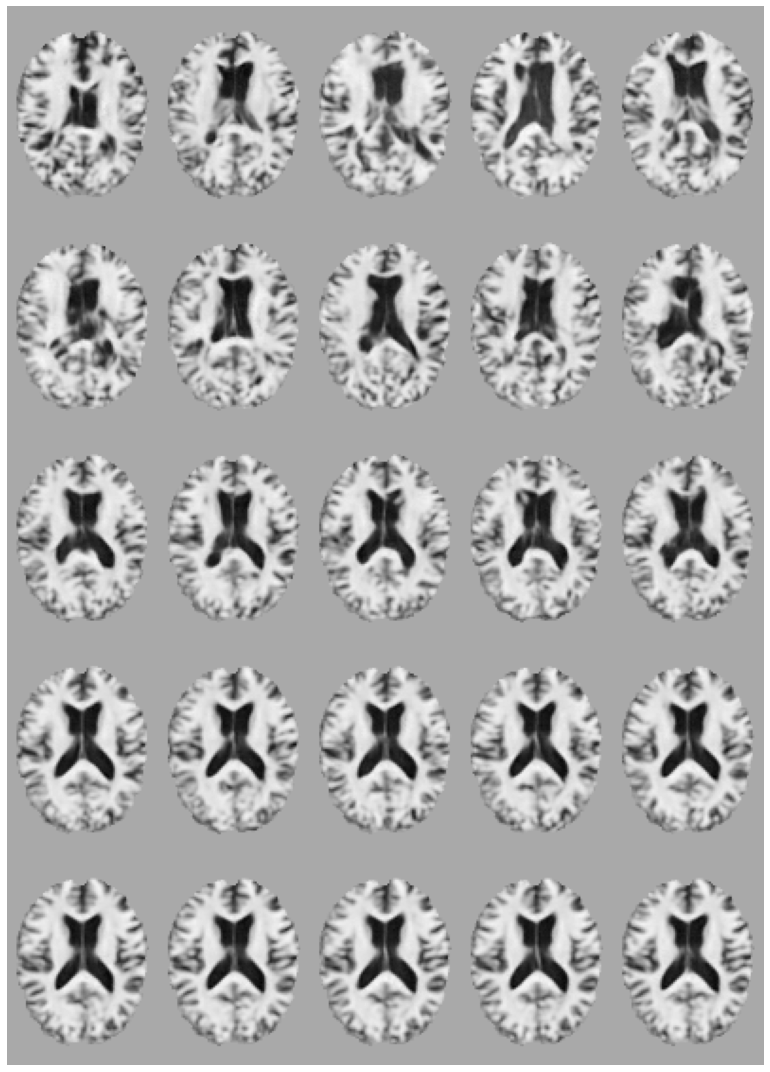}
  }
\end{figure}
\begin{figure}[p]
\floatconts
  {fig:VamPriorClusters}
  {\vspace{-1.8em}\caption{(MS) Clusters discovered by VamPrior. } \vspace{-1.8em}}
  {
     \includegraphics[trim={0 0 0 0}, width=\linewidth]{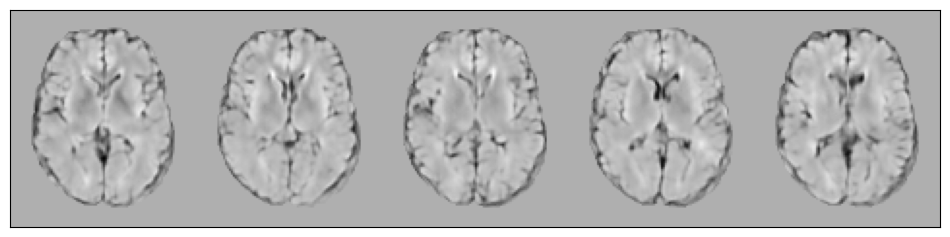}
     \includegraphics[trim={0 0 0 0}, width=\linewidth]{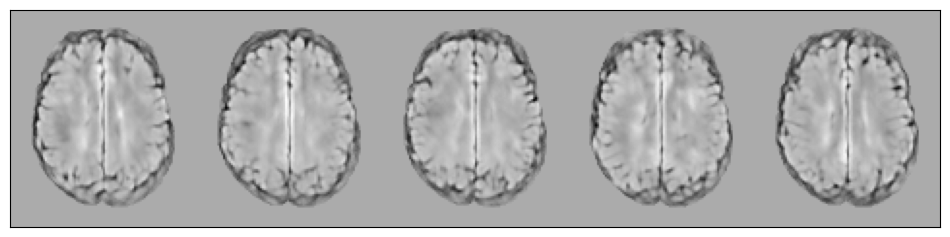}
     \includegraphics[trim={0 0 0 0}, width=\linewidth]{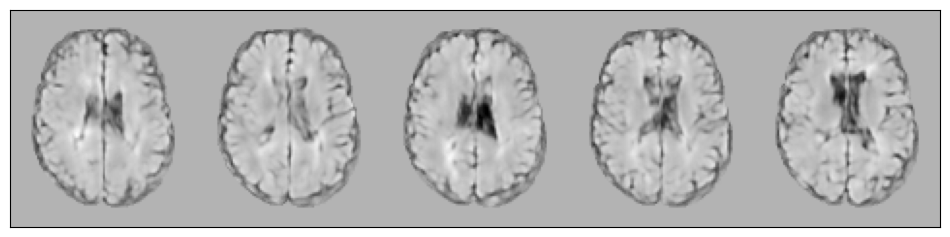}
     \includegraphics[trim={0 0 0 0}, width=\linewidth]{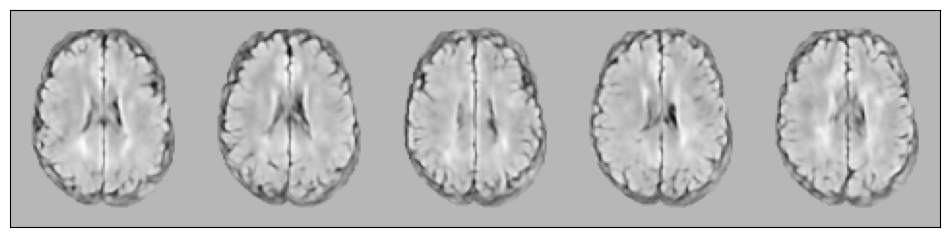}
     \includegraphics[trim={0 0 0 0}, width=\linewidth]{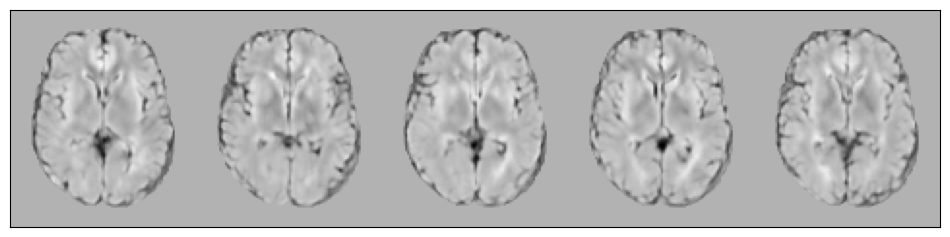}
  }
\end{figure}

\end{document}